\documentclass[letterpaper, 10 pt, conference]{ieeeconf}  

\IEEEoverridecommandlockouts                              

\usepackage{color,soul}
\usepackage{amsmath}
\usepackage{multirow}
\usepackage{graphicx}
\usepackage{booktabs}
\usepackage{lineno}
\usepackage{hyperref}

\title{\LARGE \bf
	{ARMBench}: An Object-centric Benchmark Dataset\\ for Robotic Manipulation
}

\author{Chaitanya Mitash$^{1}$, Fan Wang$^{1}$, Shiyang Lu$^{2}$, Vikedo Terhuja$^{1}$,\\ Tyler Garaas$^{1}$, Felipe Polido$^{1}$, Manikantan Nambi$^{1}$
\thanks{$^{1}$Amazon Robotics, MA, USA. \{cmitash, fanwanf, terhuja, tggaraas, polidof, mnambi\}@amazon.com}%
\thanks{$^{2}$Computer Science Department, Rutgers University, NJ, USA. shiyang.lu@rutgers.edu. Work done during a co-op at Amazon Robotics.}%
}
\begin{document}
	\maketitle

	\begin{abstract}
	This paper introduces Amazon Robotic Manipulation Benchmark ({ARMBench}), a large-scale, object-centric benchmark dataset for robotic manipulation in the context of a warehouse. Automation of operations in modern warehouses requires a robotic manipulator to deal with a wide variety of objects, unstructured storage, and dynamically changing inventory. Such settings pose challenges in perceiving the identity, physical characteristics, and state of objects during manipulation. Existing datasets for robotic manipulation consider a limited set of objects or utilize 3D models to generate synthetic scenes with limitation in capturing the variety of object properties, clutter, and interactions. We present a large-scale dataset collected in an Amazon warehouse using a robotic manipulator performing object singulation from containers with heterogeneous contents. ARMBench contains images, videos, and metadata that corresponds to 235K+ pick-and-place activities on 190K+ unique objects. The data is captured at different stages of manipulation, i.e., pre-pick, during transfer, and after placement. Benchmark tasks are proposed by virtue of high-quality annotations and baseline performance evaluation are presented on three visual perception challenges, namely 1) object segmentation in clutter, 2) object identification, and 3) defect detection. ARMBench can be accessed at \href{http://armbench.com/}{http://armbench.com}
	\end{abstract}

	\section{Introduction}
	\label{sec:introduction}
	Robotic systems for object handling in warehouses can expedite fulfillment of customer orders by automating tasks such as object picking, sorting, and packing. However, building reliable and scalable robotic systems for object manipulation in warehouses is not trivial. Modern warehouses process millions of unique objects with diverse shapes, materials, and other physical properties. These objects are often stored in unstructured configurations within containers which pose challenges for robotic perception and planning. From 2015 to 2017, the Amazon Robotics Challenge (ARC) helped push the state-of-the-art for robotic systems in a pick-and-place task representative of a warehouse \cite{colling2018progress, correll2016analysis}. Nevertheless, the competition could not incorporate challenges of large-scale operations. Fundamental research still needs to be carried out to enable visual perception algorithms such as object segmentation and identification to generalize to a wide variety of unseen objects and configurations. Additional problems (such as defect detection) and metrics (measuring uncertainty in prediction) need to be defined to capture the scale and high-precision requirements of such systems. 

\begin{figure}[t!]
	\includegraphics[width=\linewidth, keepaspectratio]{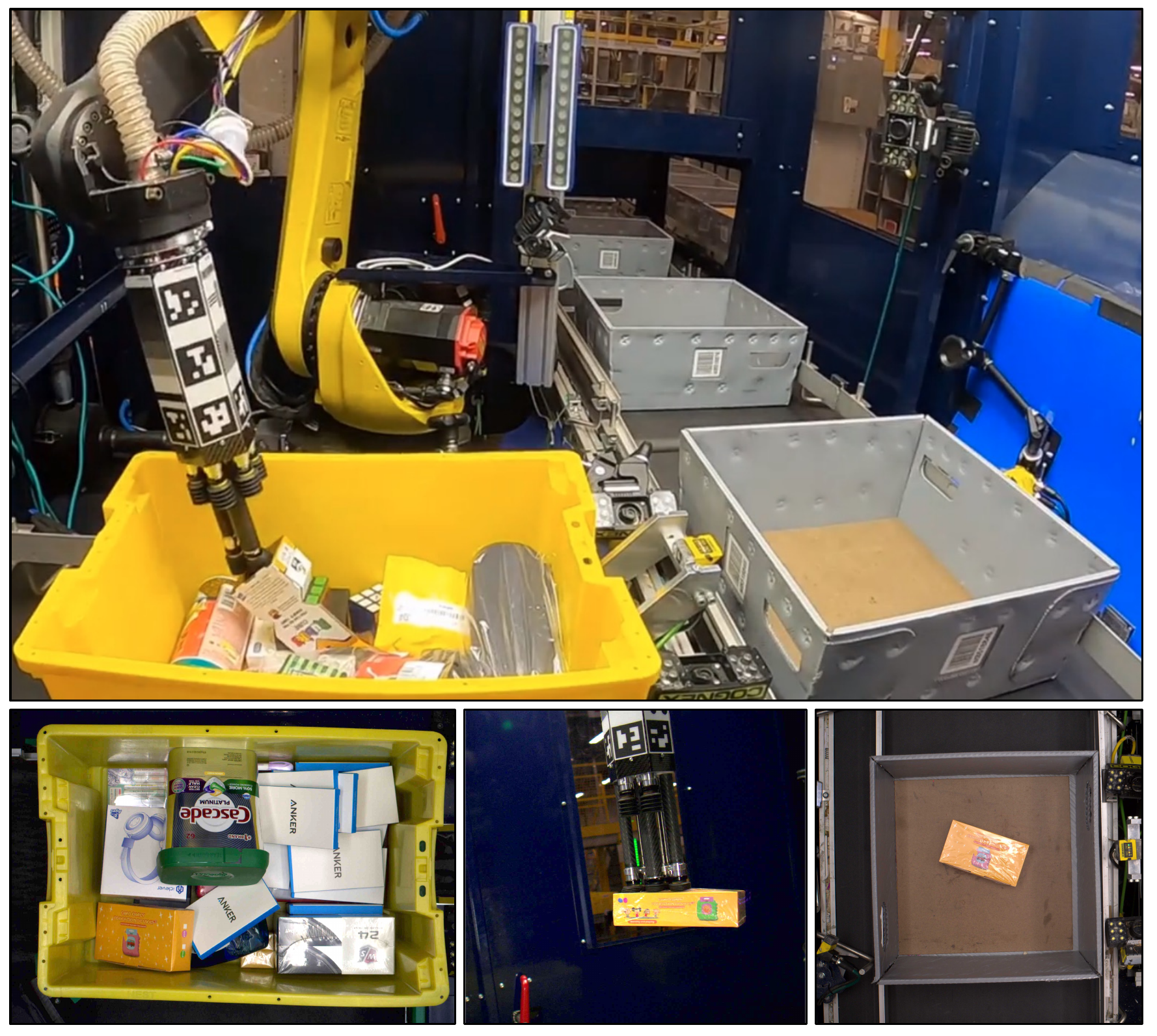}
	\caption{A large-scale object dataset is collected using a robotic manipulation system operating in an Amazon warehouse. The robotic arm picks one object at a time from the yellow container and places it in a gray tray (top). The dataset contains images for different phases of manipulation i.e., image of objects in the yellow container before picking (bottom-left), during transfer (bottom-mid) and after placement (bottom-right). In addition to sensor data, the dataset also provides high-quality annotations for tasks such as object segmentation, object identification, and defect detection.}
	\label{fig:intro}
	\vspace{-0.2in}
\end{figure}

Benchmarks and datasets such as ImageNet \cite{russakovsky2015imagenet} and MS-COCO \cite{Lin2014MicrosoftCC} have enabled significant performance improvement in computer vision tasks such as image classification and segmentation. No sizeable dataset exists that capture the desired variety of objects, configurations, and interactions in the context of robotic manipulation. Large repositories of 3D shape models \cite{calli2017yale, chang2015shapenet, collins2022abo, downs2022google} enable generating a variety of scenarios with a rich set of annotations in simulation. Nevertheless, they may fail to capture certain physical properties of objects and interactions encountered during manipulation from heterogeneous clutter.
Existing real-world datasets \cite{hodan2018bop} operate under closed set assumption with a small number of object types. Such assumptions prevent evaluating algorithms in terms of its generalization capabilities over novel objects which is critical in large scale operations. Additionally, these datasets only deal with static scenes with objects in near perfect conditions and do not consider interactions with a robotic manipulator.

In this paper, we present {ARMBench}, a large-scale benchmark dataset for a robotic pick-and-place task that captures a wide variety of warehouse objects and configurations. The dataset comprises images and videos for different stages of robotic manipulation, namely pick, transfer, and place. It includes metadata such as descriptions and reference images for objects in the container. Each pick-and-place activity is also annotated with the identity of the object being manipulated, and the outcome of the manipulation i.e., whether it was successful (a single object was picked and placed) or if it resulted in a defect. The dataset can be used to study different visual perception problems in the context of robotic manipulation. This paper provides novel benchmarks with annotations and baseline performance metrics for: 
\begin{itemize}
	\item \textbf{Object Segmentation} including 450,000+ high-quality manual labels for object segments on 50,000+ images. Variations in objects and degree of clutter present a novel challenge for instance segmentation algorithms.
	\item \textbf{Object Identification} presenting an open set object identification and confidence estimation challenge for robotic manipulation. With 190,000+ unique objects in varying configurations, the dataset will be used to benchmark image retrieval and few-shot classification methods with uncertainty estimation.
	\item \textbf{Defect Detection} with manually assigned labels for rare, but costly, robot-induced defects such as multi-object-pick and packaging defects. The dataset contains 19,000+ images and 4,000+ videos of activities with defects, and 100,000+ activities without defects.
\end{itemize}

	\section{Related Work}
	\label{sec:related_work}
	\begin{figure*}[h!]
	\includegraphics[width=\linewidth, keepaspectratio]{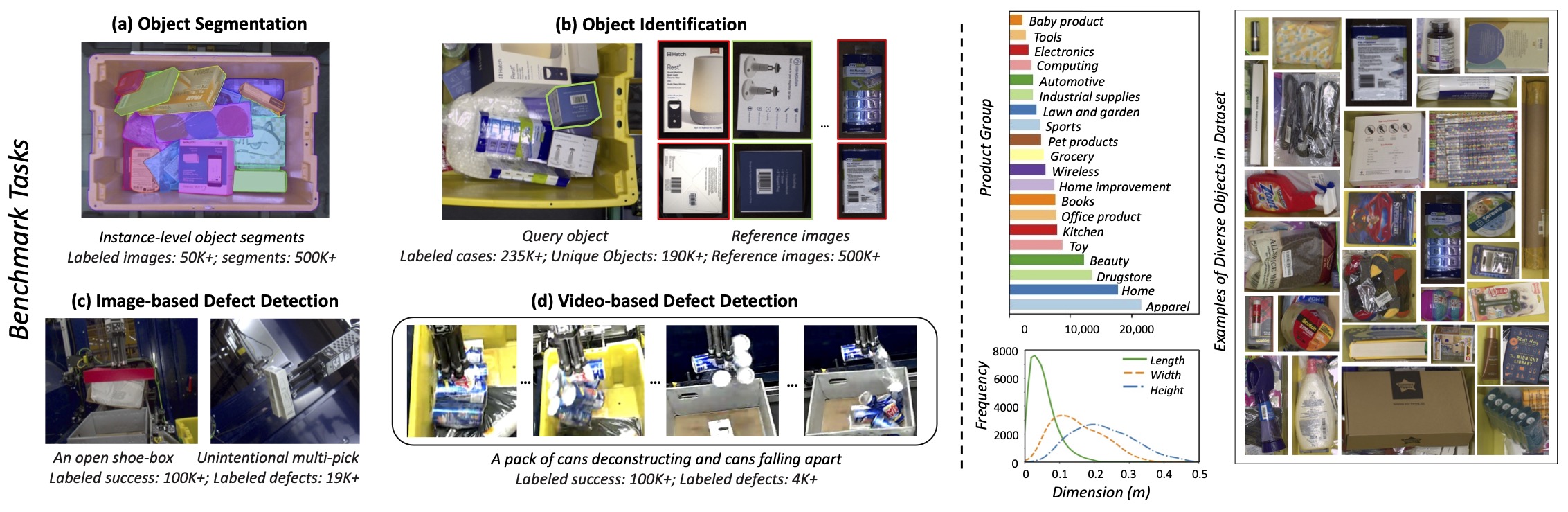}
	\caption{(left) Benchmark tasks and annotation statistics on the {ARMBench} dataset. (right) Distribution of product-groups and object dimensions for 190,00+ unique objects in the dataset.}
	\label{fig:contributions}
	\vspace{-0.1in}
\end{figure*}

\subsection{Benchmarking in Robotic Manipulation}

A recent benchmarking effort for robotic manipulation \cite{calli2021guest} considers challenges in mechanical design, grasp planning and deformable object manipulation but does not focus on the complexities of underlying perception tasks. An annual competition \cite{hodan2018bop} benchmarks performance of relevant perception algorithms such as object detection, segmentation and 6D pose estimation over a collection of datasets \cite{Brachmann2014Learning6O, Hodan2017TLESSAR, Xiang2018PoseCNNAC}. The Amazon robotics challenge \cite{Correll2018AnalysisAO, Eppner2017LessonsFT} initiated the development of other relevant datasets \cite{Rennie2016ADF, Leitner2017TheAP, zeng2017multi}. While these datasets present interesting challenges in terms of variety of configurations and large occlusions, they are limited in terms of the number of object instances with a maximum of 42 unique objects.

\subsection{Object Segmentation}

Object instance segmentation refers to simultaneously predicting pixel-level instance-mask and corresponding class labels. The technique has been widely applied in autonomous driving~\cite{zhang2016instancelevel, Brabandere2017, RenZ16}, video surveillance~\cite{gajjar2017human, Salscheider2021}, and robotics~\cite{Xie2021, Danielczuk2018, Liao2010}.  The introduction of large-scale labeled dataset such as MS-COCO~\cite{Lin2014MicrosoftCC}, PASCAL VOC~\cite{PASCAL2009}, and Cityscapes~\cite{DBLP:journals/corr/CordtsORREBFRS16} has significantly advanced the state-of-the-art in detection and segmentation, particularly for common object categories from WordNet~\cite{Fellbaum1998}. These datasets serve as a standard benchmark for evaluating computer vision models but are not representative of objects a robotic manipulator would interact with. Representative datasets such as the MVTec D2S dataset~\cite{DBLP:journals/corr/abs-1804-08292} are limited in size and diversity of objects. To obtain data at scale, a common strategy is to generate synthetic data with physics simulators and rendering tools~\cite{danielczuk2019segmenting, DBLP:journals/corr/MahlerLNLDLOG17, DBLP:journals/corr/abs-1906-07480, Xie2021, DBLP:journals/corr/abs-1809-10790}. Nevertheless, synthetic datasets are limited by the availability of high-quality 3D object models, and inherently carry a {\it sim2real} gap~\cite{ DBLP:journals/corr/abs-1809-10790}. This work introduces a real-world, large-scale dataset for object segmentation that captures a wide variety of objects and configurations relevant to robotic manipulation.

\subsection{Object Identification}

Object identification refers to the task of exactly identifying the object specified by an image segment. In an open-set setting it is often posed as an image retrieval problem i.e., given a query image and a database of candidate images, rank them according to their similarity to the query image. Various approaches have been used to tackle this problem such as aggregating pre-defined local features \cite{sivic2003video, philbin2007object, jegou2010aggregating, tolias2016image}, computing similarity metrics over features derived from large-scale image classiﬁcation training \cite{babenko2014neural, tolias2015particular, garcia2019learning}, and metric learning with pairs of matching and non-matching images \cite{radenovic2018fine, tolias2020learning}. Common benchmarks for image retrieval consider landmark datasets \cite{philbin2007object, philbin2008lost, weyand2020google} and retail datasets such as DeepFashion \cite{liu2016deepfashion, ge2019deepfashion2}, Online Products dataset \cite{oh2016deep}, RPC \cite{wei2019rpc}, RP2K \cite{peng2020rp2k}, Products-10K \cite{bai2020products}, and AliProducts \cite{cheng2020weakly}. The product images in these datasets are online store images, customer images or photos from retail stores. Alternatively, the images in the proposed dataset are representative of how objects are stored in a warehouse, with different types of packaging and in cluttered configurations. The robotic manipulation context not only provides a unique set of challenges for object identification in terms of occlusion and viewpoint variations but also imposes stringent requirements in terms of precision.

\subsection{Defect Detection}

Few image datasets exist for objects with defects. Prior research on visual defect detection has focused on surface defects for individual objects such as LED chips\cite{Lin2019}, fabrics \cite{Blanes2019}, and metals \cite{Yanqi2021}. The DAGM 2007 Competition dataset \cite{DAGM2007} comprises 6,900 synthetic images with six different types of surface defects. The MVTec Anomaly Detection dataset comprises 5,354 color images corresponding to 15 object and texture categories with 70 different types of defects such as scratches, dents, contaminations, and structural changes \cite{Bergmann2019}. This is the first dataset to capture defects in the context of robotic manipulation. 

Datasets for defect detection in videos primarily focus on anomaly detection methods for events such as throwing objects, loitering, running \cite{Liu2018}, crowded scenes \cite{Li2013}, and anomalous pedestrian patterns \cite{Landi2019}. These datasets contain a limited number of videos (10-50) per activity. Video classification datasets exists in the domain of sports activities \cite{Karpathy2014}, human actions \cite{Kay2017}, and holistic video understanding \cite{diba2020large}. Similar to existing research on using videos for understanding context and actions, videos can be used to understand events in robotic manipulation process such as successful and defective activities. There exists no large-scale video datasets for this purpose.

	\section{ARMBench Dataset}
	\label{sec:dataset}
	The ARMBench dataset presents: 1) a collection of sensor data acquired by a robotic manipulation workcell performing pick-and-place operation, 2) metadata and reference images for objects in containers, 3) a set of annotations acquired either automatically, by virtue of the system design, or via manual labeling, and 4) tasks and metrics to benchmark perception algorithms for robotic manipulation. Fig.\ \ref{fig:contributions} illustrates the benchmark tasks and variety of objects captured in the dataset. The dataset captures diversity in objects with respect to Amazon product categories as well as physical characteristics such as size, shape, material, deformability, appearance, fragility, etc. 

The data collection platform is a robotic manipulation workcell performing pick-and-place operation in a warehouse \cite{Sparrow2022}. The workcell contains a robotic arm mounted with a vacuum-based end-effector. It is presented with a heterogeneous collection of objects placed in unstructured configurations within a container (storage tote). The robotic arm is tasked with picking one object at a time (singulation) and place it on moving trays until the container is empty. The empty container ejects the workcell and is replaced by a new container. While the operation is completely autonomous, it includes a human-in-the-loop to monitor the status of each pick-and-place activity, annotate, and resolve any defects during manipulation. Multiple imaging sensors are placed in the workcell to facilitate and validate the pick-and-place operation. Following is a list of sensor data (Fig.\ \ref{fig:intro}) associated with each pick activity:
\begin{itemize}
\item Pick-image: A 5\,MP camera is used to capture a top-down image of the container.
\item Transfer-images: Multiple 5\,MP cameras are placed on different sides in the workcell to capture the moving object from different viewpoints.
\item Place-image: A top-down view of the object is captured once it is placed on the tray.
\item Video: A camera is mounted to capture 720p videos of pick-and-place manipulation processes at 30\,FPS
\end{itemize}
Additionally, the following metadata (Fig.\ \ref{fig:contributions} (b)) is available by virtue of a warehouse tracking system:
\begin{itemize}
\item Container-manifest: A list of objects present in the container along with data such as product description, coarse dimensions, and weight.
\item Reference images: One or more images of objects from previous operations within the warehouse.
\end{itemize}
The sensor data and metadata were consumed by perception algorithms required to autonomously operate the robotic workcell. Benchmarking against these algorithms would not only optimize a manipulation task such as the one used for data collection but also enable more complex and intentional manipulation. This work considers a subset of such perception tasks namely object segmentation, object identification, and defect detection. These are critical not only to make informed grasping and motion decisions but also to track the state of the objects and containers within the warehouse. The following sections will describe these tasks and present the challenges using annotations, baseline algorithms, and evaluation metrics.

	\section{Object Segmentation}
	\label{sec:segmentation}
\begin{figure}
	\centering
	\includegraphics[width = 0.5\textwidth]{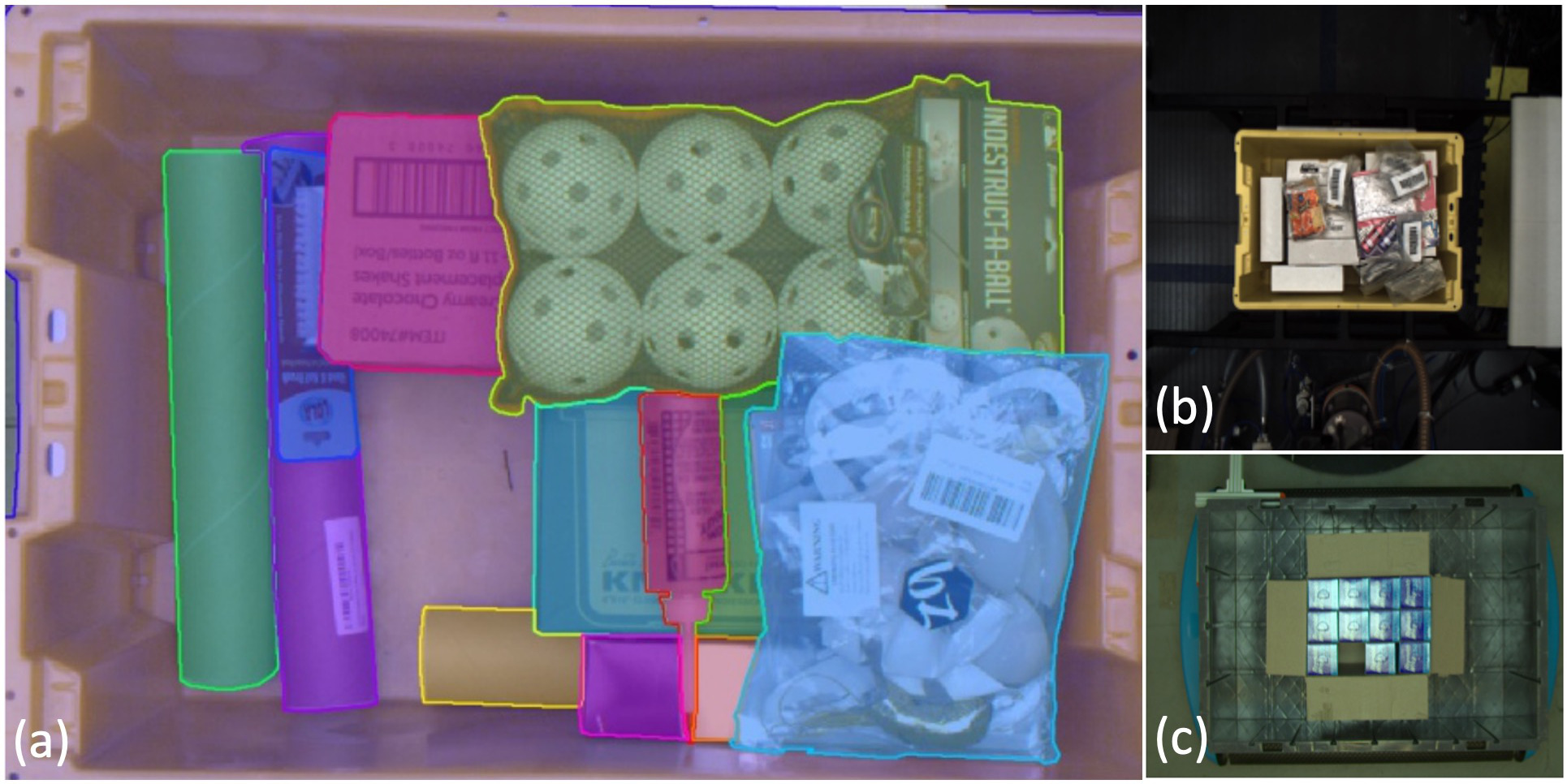}
	\caption{(a) Segmentation annotation overlaid on an image from {\it mix-object-tote}. Each identifiable item is segmented regardless of its size and occlusion. Multiple objects in the same package are considered as one object and is delineated by the boundary of the package. In particular, items wrapped in transparent packaging are segmented by the peripheral of the package, although other products may be seen through them. (b-c) Example images from {\it zoomed-out-tote-transfer-set} and {\it same-object-transfer-set} subsets representing variations in background, scale, and clutter.}
	\label{fig:segmentation_subsets}
	\vspace{-0.2in}
\end{figure}

	The object instance segmentation task is to identify and delineate distinct objects stored in containers in a warehouse. In the context of robotic object manipulation, instance segmentation is used to inform downstream robotic processes such as grasp generation, motion planning, and placement. Accuracy of instance segmentation can have an impact on picking success, object identification, and defects introduced in the process. For example, under-segmentation can result in picking multiple objects at a time, while over-segmentation can result in a bad choice of grasp leading to damage or dropping of objects. Fig.\ \ref{fig:segmentation_subsets}(a) shows manually annotated object segments on the pick-image. Presence of deformable and transparent objects in clutter makes the task challenging.

Our object instance segmentation dataset contains 50K+ images of objects stored in containers in a warehouse with 500K+ annotations. The annotations include instance-level segmentation masks and bounding box for two classes (object and container). Technicians with task-specific training generated high-quality annotations for object boundaries and object class which are verified by two additional quality assurance technicians.

We divide the object segmentation dataset into three subsets. The primary set, \textit{mix-object-tote}, comprises 44,253 images and 467,225 annotations of objects in yellow and blue storage totes. The totes contain a heterogeneous clutter of objects with an average of 10.5 object segments (ranging from 1 to 50 segments) in each image. The other two subsets, namely \textit{zoomed-out-tote-transfer-set} and \textit{same-object-transfer-set} (Fig.\ \ref{fig:segmentation_subsets}(b) and (c)) enable us to understand the impact of variation in data distribution. The \textit{zoomed-out-tote-transfer-set} subset with 5,837 images and 43,401 annotations captures images of containers from a different warehouse. It poses a transfer learning challenge due to significant differences in background, scale, and object distribution. The \textit{same-object-transfer-set} subset contains 3,323 images and 12,664 annotations. It captures a common and visually challenging scenario in warehouses where multiple instances of the same object are tightly packed in a container. 


To establish a performance baseline, we trained Matterport's implementation of Mask R-CNN~\cite{matterport_maskrcnn_2017, DBLP:journals/corr/HeGDG17} with ResNet-50 backbone~\cite{DBLP:journals/corr/HeZRS15} on the {\it mix-object-tote} dataset. Default training schedule (for MS-COCO) and hyper-parameters were used along with a train-valid-test split of 0.7:0.15:0.15. 
Table~\ref{table:segmentation_results_inference} shows the results for our baseline experiment. Mean average precision ($mAP$) for a threshold of 0.5 ($mAP_{50}$) and 0.75 ($mAP_{75}$) are used to evaluate the performance of the baseline model on test set. 

We observe that applying model weights trained on {\it mix-object-tote} to the {\it zoomed-out-tote-transfer-set} ($mAP_{50}=0.25$) and {\it same-object-transfer-set} subsets ($mAP_{50}=0.11$) yields poor results. While techniques like transfer learning can improve performance on a new scenario when a reasonable amount of domain-specific labeled data is available, labeling specifically for each variation is time-consuming, if feasible at all. The ultimate goal is to readily transfer segmentation to new scenarios with minimal additional annotations.


\setlength{\tabcolsep}{4pt}
\begin{table}
\centering
\caption{Mask R-CNN performance for object segmentation task. The model was trained on \textit{mix-object-tote} dataset}
\label{table:segmentation_results_inference}
\begin{tabular}{@{}rccc@{}}
    \hline
 & mix-object-tote & zoomed-out-tote- & same-object- \\ 
 & & transfer-set & transfer-set \\ \hline
$mAP_{50}$  & 0.72 & 0.25 & 0.11 \\ \hline
$mAP_{75}$  & 0.61 & 0.19 & 0.10\\ \hline
\end{tabular}
\end{table}
\setlength{\tabcolsep}{1.4pt}

\begin{figure}[h]
	\centering
	\includegraphics[width = 0.45\textwidth]{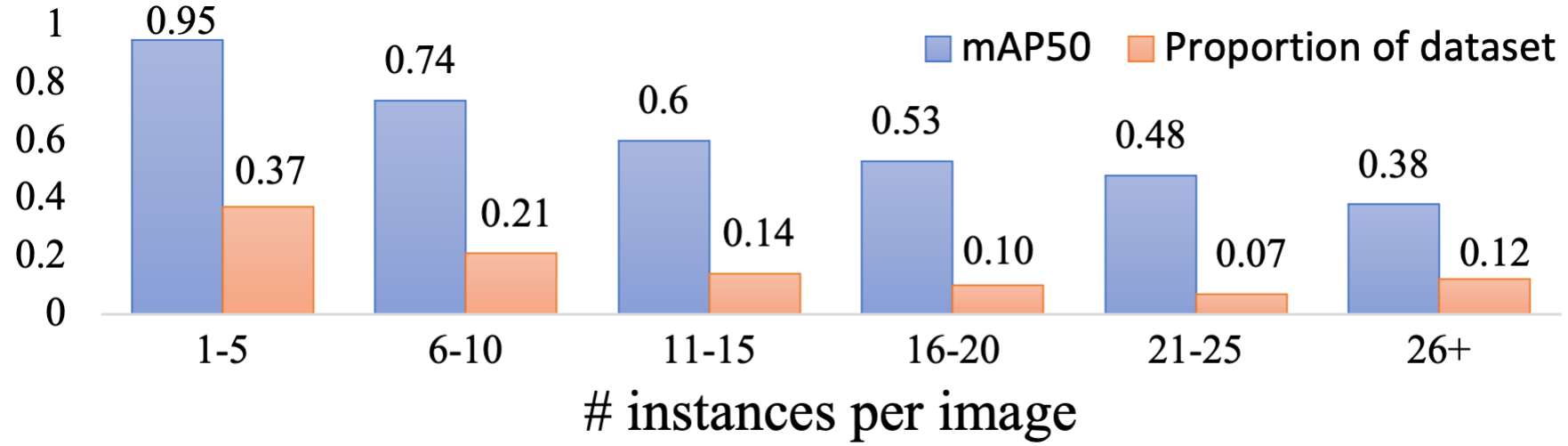}
	\caption{Performance on {\it mix-object-tote} with varying degree of clutter.}
	\label{fig:segmentation_clutter}
	\vspace{-0.15in}
\end{figure}

We observe that segmentation performance for our baseline model has a strong correlation to the level of clutter. Fig.~\ref{fig:segmentation_clutter} shows that the performance drops significantly as the number of ground-truth object instances increases in the image. The $mAP_{50}$ score drops sharply from 0.95 when the tote has one to five object instances to a low of 0.38 when there are more than 26 object instances in the image. This motivates developing algorithms that are robust against clutter and occlusion to further improve object segmentation performance.

	\section{Object Identification}
	\label{sec:identification}

Object identification (ID) is the task of exactly identifying an image segment as one of the objects within a database. In the robotic manipulation context, this task is applicable both before and after picking the object. In the pre-pick stage, identifying an object segment within the tote allows accessing any stored models or attributes of the object from past experience which can be used for manipulation planning purposes. In the post-pick stage, ID has access to the segment of the object being manipulated both within the tote as well as when it is attached to the robotic arm. Accurately identifying which object is being transferred from one container to another is critical to tracking the object within a warehouse, thereby maintaining a container-manifest. This also allows posing ID as an image retrieval challenge. The pick and transfer images with segments of the target object (acquired using an instance segmentation algorithm) are treated as {\it query images}. The reference images for all objects in the container manifest are treated as {\it gallery} images. The challenge is to compare the {\it query images} to {\it gallery images} to find a match.

The benchmark dataset for this task contains 235K+ labeled pick activities corresponding to 190K+ unique objects. Each pick activity comprises one query image from the pick scene and up to three query images from the transfer phase. A ground-truth ID annotation is automatically acquired using multiple barcode scanners that are placed in the workcell. In cases where no barcode is scanned, a human operator manually scans the barcode of the object after it ejects the workcell. Pick activities are accompanied by a set of reference images for objects in the container manifest. These images are captures of the object from previous operations within the warehouse. While up to six reference images are sampled per object, reference images are not available for some objects. Such cases are representative of scenarios when a new object is introduced into the warehouse. To tackle such cases, an ID algorithm needs to model some notion of confidence and prevent false-positive prediction. A {\it test} set is sampled for evaluating baseline algorithms on the dataset. This set contains 50,000 pick activities where at least one reference image is available for the picked object. Another set ({\it test-uncertainty}) is derived from the {\it test} set by ignoring reference images of the picked object for 20\% of the cases. This set is used to evaluate the behavior of ID on novel objects coming into the warehouse.

\setlength{\tabcolsep}{6pt}
\begin{table}[h]
	\centering
	\caption{Evaluating Top-k Object Retrieval}
	\label{table:id_results}
	\begin{tabular}{c|c|c|c}
		\hline
		recall@k (pre/post-pick) & k=1 & k=2 & k=3\\
		\hline
		ResNet50-RMAC \cite{tolias2015particular} & 71.7 / 72.2 & 81.9 / 82.9 & 87.2 / 88.2\\
		DINO-ViTS \cite{caron2021emerging} & 77.2 / 79.5 & 87.3 / 89.4 & 91.6 / 93.5\\
		\hline
		\multicolumn{4}{c}{test-uncertainty} \\
		\hline
		ResNet50-RMAC \cite{tolias2015particular} & 57.7 / 58.0 & 65.7 / 66.5 & 70.2 / 70.7\\
		DINO-ViTS \cite{caron2021emerging} & 61.9 / 63.6 & 69.9 / 71.6 & 73.3 / 74.8\\
		\hline
	\end{tabular}
\end{table}
\setlength{\tabcolsep}{1.4pt}

\begin{figure}[t!]
	\includegraphics[width=\linewidth, keepaspectratio]{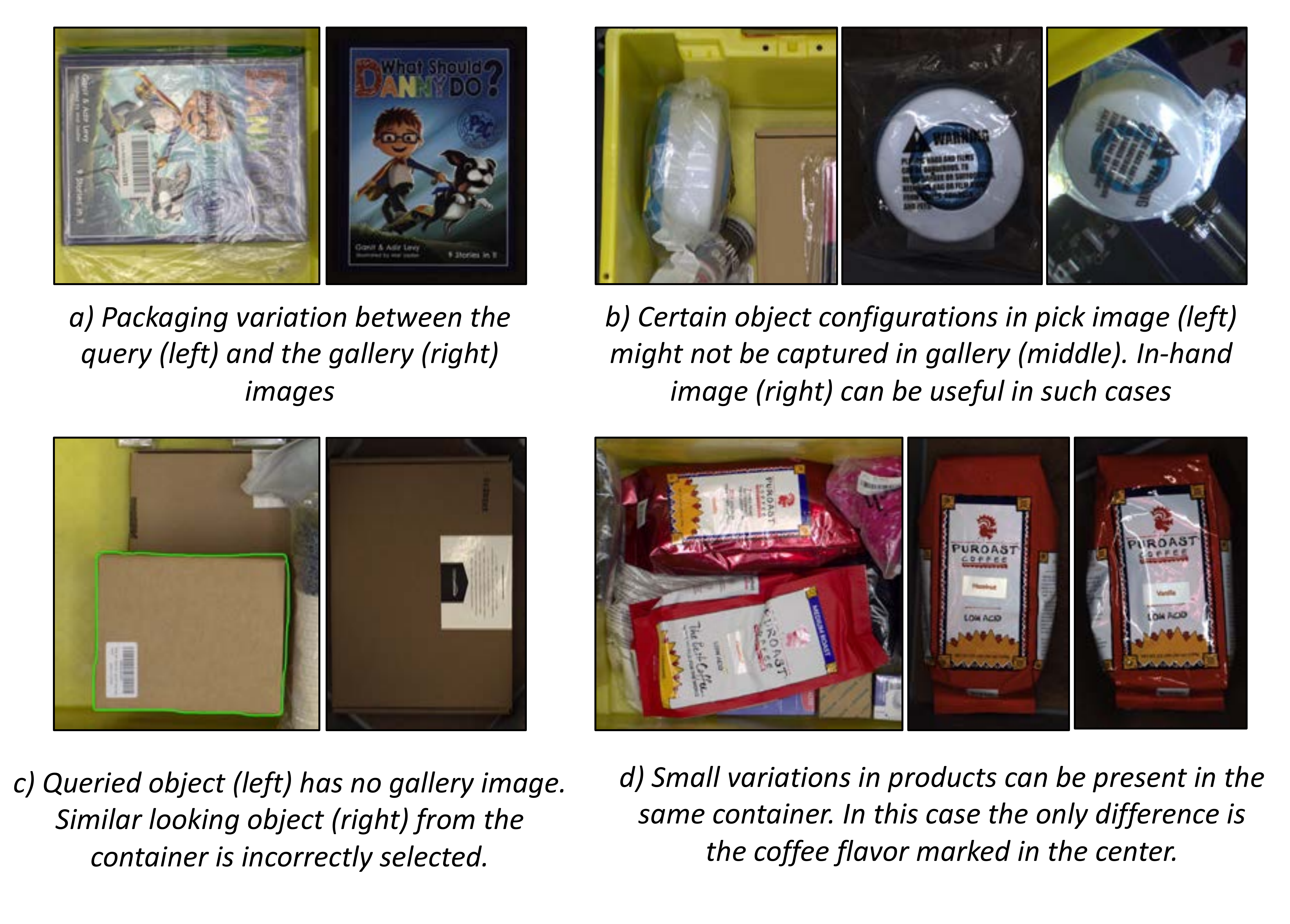}
	\vspace{-0.2in}
	\caption{Challenging cases for Object Identification}
	\label{fig:id-cases}
\end{figure}

\begin{figure}[t]
	\includegraphics[width=\linewidth, keepaspectratio]{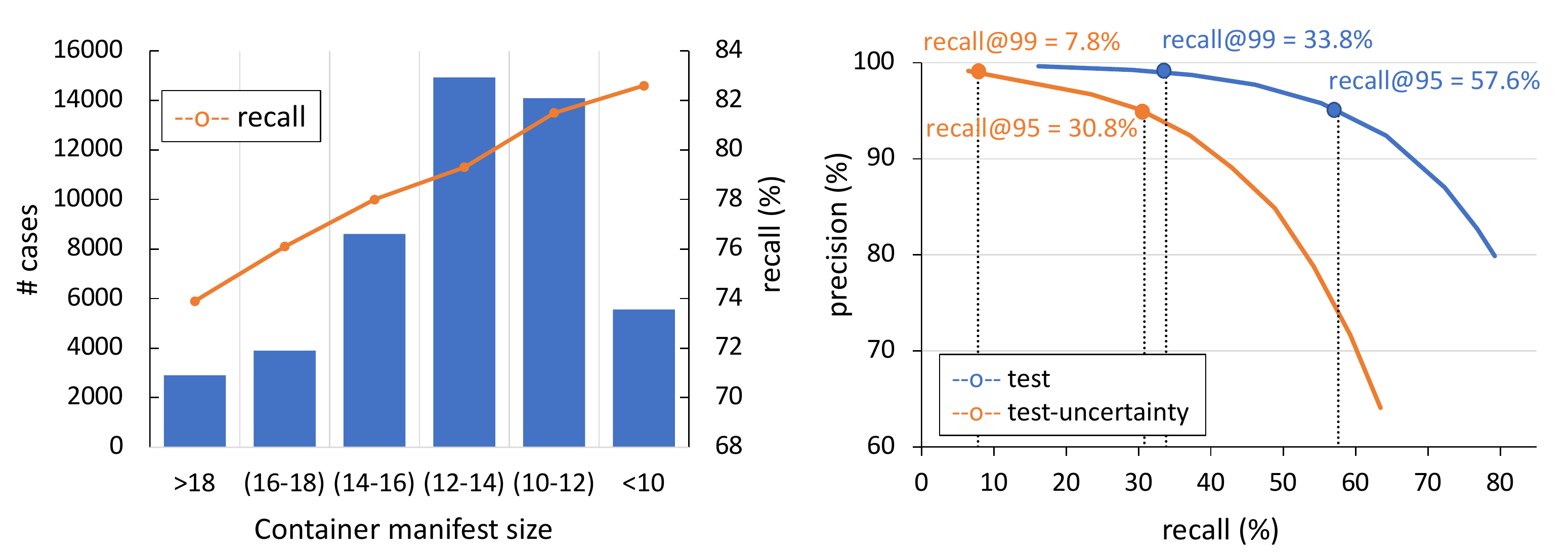}
	\caption{Container manifest size (left) is indicative of the number of images that the algorithm needs to select from. The precision-recall curve (right) shows the need for a confidence model to prevent false-positive predictions.}
	\label{fig:id-plots}
\end{figure}

 Table\ \ref{table:id_results} shows results of object retrieval with baseline algorithms on the two sets. For the first baseline, a 512d image descriptor is extracted from a ResNet50 backbone via aggregating features \cite{tolias2015particular} pre-trained for classification on ImageNet dataset \cite{russakovsky2015imagenet}. The second baseline utilizes a 384d feature vector pre-trained via self-supervision \cite{caron2021emerging} on a vision transformer. A cosine similarity is computed between feature embeddings for {\it query} and {\it gallery} images to get the closest match. Evaluation is performed both over pre-pick (pick image only) and post-pick (pick and transfer images) scenarios. Although transfer images are significantly different in terms of presentation to reference images, they provide multiple views of the object which improves the overall retrieval rate. Fig.\ \ref{fig:id-cases} shows some of the challenges associated with ID on this dataset. Large variations in appearance for the same object and the similarities between different objects makes the dataset challenging. The challenge increases with the size of container manifest as seen in Fig.\ \ref{fig:id-plots}(left). Fig.\ \ref{fig:id-plots}(right) shows the precision-recall curve obtained based on a rank-ratio confidence metric computed as $(1-\frac{c_2}{c_1})$, where $c_1, c_2$ are softmax probabilities corresponding to the first and second ranked objects. The plot highlights recall rates at high precision values as mis-identifications can lead to costly scenarios, such as an object getting lost within the warehouse. While methods like contrastive learning over the training set can improve the top-1 retrieval rate, achieving a high recall rate within the precision constraints would require methods to perform uncertainty estimation and leverage additional modalities such as text and dimensions. 

	\section{Defect Detection}
	\label{sec:defect_detection}

\begin{figure}[t]
	\centering
	\includegraphics[width=0.5\textwidth]{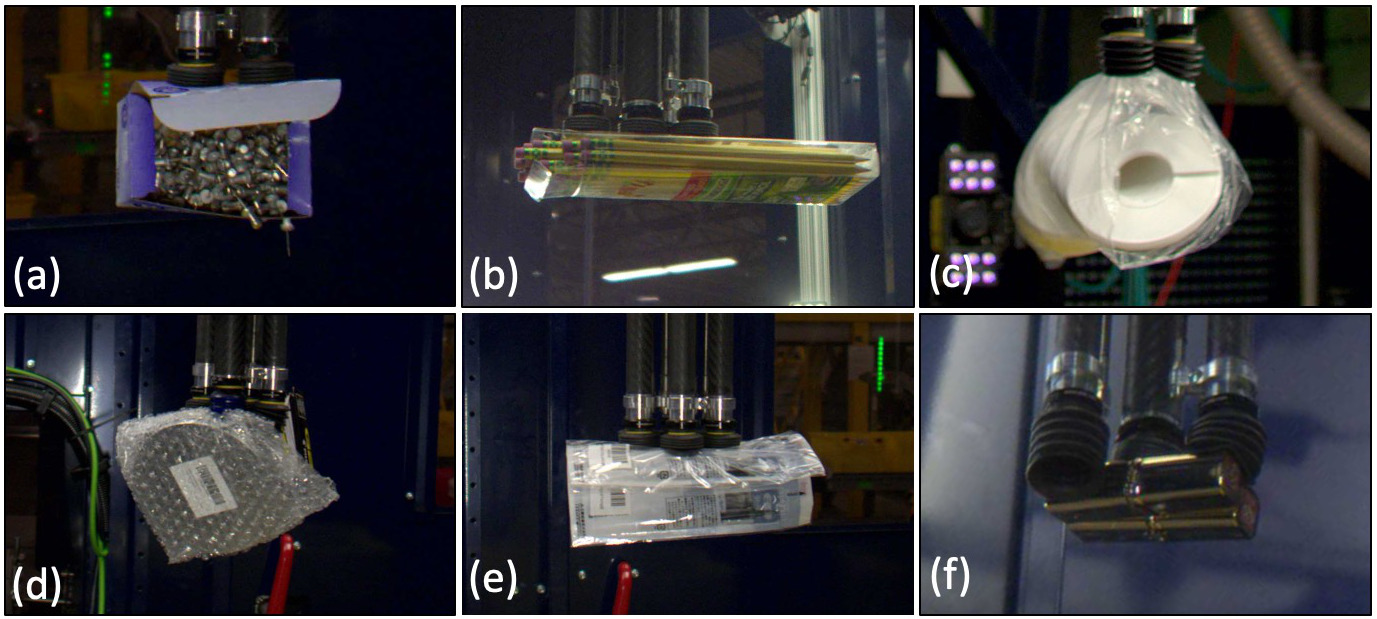} 
	\caption{ Multi-view images in the defect detection dataset showing (a)--(c) \textit{package-defect} and (d)--(f) \textit{multi-pick} defect for different types of objects. }
	\label{fig:defects}
	\vspace{-0.1in}
\end{figure}

The defect detection task is to identify if a robotic manipulation activity resulted in a defect. Two types of robot-induced defects are included in the dataset: 1) \textit{multi-pick}, and 2) \textit{package-defect}. \textit{Multi-pick} is used to describe activities where multiple objects were picked and transferred from the source container to the destination container. \textit{Package-defect} is used to describe activities where the object packaging \textit{open}ed and/or the object separated into multiple parts (\textit{deconstruction)}. Two subclasses, \textit{open} and \textit{deconstruction}, are defined for package-defect.  
Fig.\ \ref{fig:defects} shows examples of multi-pick and package-defect in our dataset. Multi-pick are often observed when there is a high degree of clutter, there are multiple instances of the same object, or when objects of significantly different sizes are placed together. Fig.\ \ref{fig:defects} (a-c) shows package-defect on a variety of objects. Defects on deformable objects like plastic bags can be challenging for visual detection. 

Our dataset comprises 19,303 images of objects from multiple viewpoints (Transfer-images) and 4,070 videos of pick-and-place activities that resulted in a defect. 
Videos are excluded from our dataset for multi-pick defect as such defects are not observable along specific viewpoints. Multi-view Transfer-images are best suited to detect multi-pick defect. On the other hand, \textit{open} and \textit{deconstruction} defects can happen at any time during an activity. As a result, they are best captured using videos. The dataset includes 100,000 images of objects and videos of activities that do not have any defects and are defined as \textit{nominal} activities. 
Tables\ \ref{tab:defect_image_baseline} and \ref{tab:defect_video_baseline} shows the distribution of defect types in our dataset. In addition to Transfer-images, the dataset includes Pick-image and Place-image that provide context for an activity.

A two-step process was used to annotate data. A technician operating our system labeled each activity as successful/nominal (a single object transferred from the source container to the destination container), \textit{multi-pick}, \textit{open}, or \textit{deconstruction} defect. Expert annotators verified the annotations for each activity and augmented the annotations for Transfer-images as multi-pick, or package-defect if a defect was observable, and as {nominal} if no defect was observable in the image. In addition to the defect type, we also provide segmentation polygons for the objects to enable development of models that can benefit from additional attention cues. For video annotations, expert annotators verified the type of package defect, i.e., \textit{open} and \textit{deconstruction}, observed in the video. In addition to the type of defect observed in each video, the index of the first frame where a defect becomes observable is also provided to enable development of real-time defect detection methods.

To establish a baseline for defect detection, we performed two experiments. In the first experiment, we train an image classifier with ResNet-50 \cite{DBLP:journals/corr/HeZRS15} backbone, global average pooling, and focal loss for predicting the type of defect observed in the Transfer-images. In the second experiment, we trained a multi-scale vision transformer model (MViT-B) \cite{fan2021MultiscaleVT} for action classification on videos. Since a defect can be introduced at any time during the manipulation process, we uniformly sampled 32 frames ($\sim$5\,FPS) from each video for training. 
The classification head outputs a two-channel vector that predicts binary classification on two categories: \textit{open} and \textit{deconstruction}. 
We used a train-test split of 0.7:0.3 for multi-pick and package-defects. The nominal category in the train set was downsampled to match the size of the defect category to compensate for class imbalance. 10,000 samples from the nominal category were added to our test set. 

Table\ \ref{tab:defect_image_baseline} and \ref{tab:defect_video_baseline} show performance of baseline models for defect detection on images and videos. We used recall and false positive rate (fpr) as metrics to evaluate performance over defect classes. A missed defect (lower recall) is more expensive than classifying a nominal activity as defective (fpr). Results for image defect detection shows that multi-picks are a harder to detect than package-defects. On the other hand, results for video defect detection show that open defects are harder to detect than deconstruction. There is significant scope for improvement in defect detection methods to be effective in warehouses operations which typically require high recall ($>$0.95) and low fpr ($<$0.01). 

\begin{table}[t]
	\centering
	\caption{Baseline for single-view image defect detection}
	\label{tab:defect_image_baseline}
	\setlength{\tabcolsep}{3pt}
	\begin{tabular}{c|c||c|c||c}
	    \hline 
	    {model} & {metric}  &  {multi-pick}  & { package-defect} & combined \\
	    \hline  
		\multirow{3}{*}{ResNet-50 \cite{DBLP:journals/corr/HeZRS15}}   & count     &  7,813  & 11,490  & 19,303 \\& recall     &  0.34  & 0.73  & 0.57 \\ 
									                	               & fpr        &  0.05  & 0.05  & 0.05  \\  \hline
	\end{tabular}
\end{table}
\setlength{\tabcolsep}{1.4pt}

\begin{table}[t]
	\centering
	\caption{Baseline for video defect detection}
	\label{tab:defect_video_baseline}
	\setlength{\tabcolsep}{3pt}
	\begin{tabular}{c|c||c|c||c}
	    \hline 
		model & metric  & open & deconstruction & combined \\ 
		\hline
		\multirow{3}{*}{MViT-B \cite{fan2021MultiscaleVT}}   
		& count     &  2,951  &     2,165  &   4,070 \\
		& recall    &  0.69   &     0.79   &   0.73 \\ 
	    & fpr 	    &  0.23   &     0.03   &   0.13 \\ \hline
	\end{tabular}
\end{table}
\setlength{\tabcolsep}{1.4pt}

	\section{Discussion and Future Work}
	\label{sec:discussion}
	In this work we introduced ARMBench, a large-scale, object-centric benchmark dataset for robotic manipulation in warehouses. The object segmentation benchmark presents challenges relating to clutter, deformable and transparent packaging as well as the problem of degrading performance with different backgrounds and storage configurations. The identification benchmark presents an open set recognition challenge on a wide variety of objects. Additionally, images of the same object can vary significantly due to differences in configurations and packaging variations while images of two different objects can appear similar. Missing reference images and high precision requirement makes the benchmark well suited to evaluate uncertainty estimation algorithms.
Finally, the defect detection benchmark presents a unique set of challenges such as detection of multi-pick, opening, and deconstruction of packages. Annotations and baselines are provided both for a single-shot as well as video-based detection of such events. 

Our intention is for this dataset to grow over time with a goal to increase the number of unique objects, environments, and benchmark tasks. Large-scale sensor data and fine-grained attributes of objects will enable learning generalizable representations that could transfer to other visual perception tasks. We further plan to enrich our dataset with 3D data and annotations, and propose new benchmark tasks. 
	
	\section{Acknowledgements}
	We would like to thank the Sparrow \cite{Sparrow2022} team members for deployment and operation of the robotic workcell, Aalekh (Raj) Ray Chaudhury and the Go-AI team for data annotation support and the Item-matrix team for curating the reference image dataset. We would also like to thank Joey Durham, Andy Marchese, Clay Flannigan, Parris Wellman, Jane Shi and Kapil Katyal for their valuable feedback.
	
	\bibliographystyle{IEEETranS}
	\bibliography{bibtex}	
\end{document}